\documentclass[10pt,twocolumn,letterpaper]{article}

\usepackage{wacv}
\usepackage{times}
\usepackage{epsfig}
\usepackage{graphicx}
\usepackage{amsmath}
\usepackage{amssymb}
\usepackage{wrapfig}
\usepackage{color}
\definecolor{brown}{rgb}{0.65, 0.16, 0.16}

\def\wacvPaperID{577} 

\wacvfinalcopy 

\ifwacvfinal
\def\assignedStartPage{1} 
\fi

\ifwacvfinal
\usepackage[breaklinks=true,bookmarks=false]{hyperref}
\else
\usepackage[pagebackref=true,breaklinks=true,colorlinks,bookmarks=false]{hyperref}
\fi

\ifwacvfinal
\else
\fi

\begin{document}

\title{Equine Pain Behavior Classification via Self-Supervised Disentangled Pose Representation}

\author{Maheen Rashid$^{1,2}$ ~~~~ Sofia Broomé$^3$ ~~~~ Katrina Ask$^4$ ~~~~ Elin Hernlund$^4$ ~~~~ Pia Haubro Andersen$^4$ \\ Hedvig Kjellström$^{3,5}$ ~~~~ Yong Jae Lee$^{2,6}$\\
$^1$ Univrses AB, Sweden {\tt maheen.rashid@univrses.com} ~~~~
$^2$ UC Davis, USA\\
$^3$ KTH Royal Institute of Technology, Sweden {\tt sbroome,hedvig@kth.se} \\
$^4$ SLU, Sweden {\tt katrina.ask,elin.hernlund,pia.haubro.andersen@slu.se} \\
$^5$ Silo AI, Sweden ~~~~ $^6$ UW Madison, USA  {\tt yongjaelee@cs.wisc.edu}}

\maketitle
\ifwacvfinal\thispagestyle{empty}\fi

\begin{abstract}
    Timely detection of horse pain is important for equine welfare. Horses express pain through their facial and body behavior, but may hide signs of pain from unfamiliar human observers. In addition, collecting visual data with detailed annotation of horse behavior and pain state is both cumbersome and not scalable. Consequently, a pragmatic equine pain classification system would use video of the \emph{unobserved} horse and weak labels. This paper proposes such a method for equine pain classification by using multi-view surveillance video footage of unobserved horses with induced orthopaedic pain, with temporally sparse video level pain labels. To ensure that pain is learned from horse body language alone, we first train a self-supervised generative model to disentangle horse pose from its appearance and background before using the disentangled horse pose latent representation for pain classification. To make best use of the pain labels, we develop a novel loss that formulates pain classification as a multi-instance learning problem. Our method achieves pain classification accuracy better than human expert performance with 60\% accuracy. The learned latent horse pose representation is shown to be viewpoint covariant, and disentangled from horse appearance. Qualitative analysis of pain classified segments shows correspondence between the pain symptoms identified by our model, and equine pain scales used in veterinary practice.
\end{abstract}
\section{Introduction}
\begin{figure}[ht]
    \centering
    \includegraphics[width=0.9\linewidth]{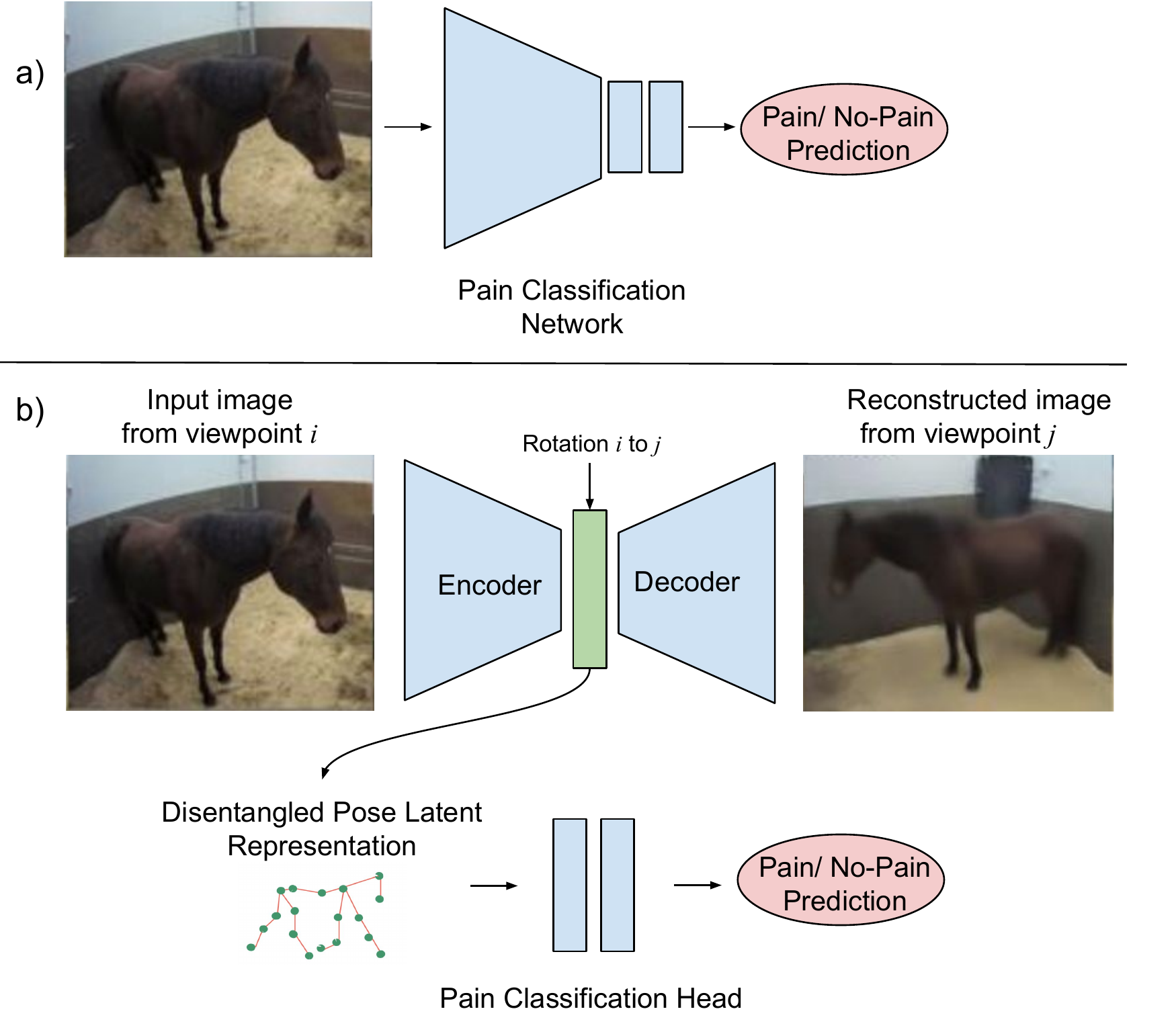}
    \caption{\textbf{Main Idea} (a) Directly training a pain classifier on video frames can result in a model that is not interpretable, and would overfit to limited training data. (b) Our method first uses self-supervised multi-view synthesis to learn a latent horse pose representation, that is then used to learn a light weight pain classifier.}
    \label{fig_egp_key_idea}
\end{figure}
Timely detection of pain in horses is important for their welfare, and for the early diagnosis and treatment of underlying disease. While horses express pain through changes in facial and body behavior, equine pain classification is a challenging problem, with expert human performance on video data at just 58\% accuracy for pain or no pain classification for acute pain~\cite{broome2019dynamics}, and 51\% accuracy for orthopaedic pain~\cite{broome2021sharing}. While self-evaluation can be used as a gold standard for determining pain in human subjects, horses, being non-verbal, lack a gold standard for pain~\cite{andersen2018can}. In addition, as prey animals, horses may hide signs of pain from human observers~\cite{coles2016no}. It is therefore difficult to ascertain if a horse is experiencing and expressing pain. 

Consequently, even though pain scales relying on facial and body behaviors are used in veterinary practice~\cite{raekallio1997comparison,price2003preliminary,sellon2004effects,graubner2011clinical,gleerup2016recognition}, determining the pain level of horses remains challenging. Furthermore, outside of an animal hospital or clinic, horse owners may underestimate the prevalence of disease, and thereby impede contact with the veterinarian~\cite{ireland2012comparison}. A computer vision system capable of determining horse pain, therefore, has great potential for improving animal welfare by enabling round the clock classification of pain and consequent diagnosis of illness. 

Human pain datasets with clear closeup facial footage, and detailed Facial Action Coding System (FACS)~\cite{ekman2002facial} annotation have been used for training human pain classification systems~\cite{lucey2011painful}. However, a similar approach is not practical for equine pain classification. First, obtaining similar annotation is also costly and time consuming for horses as it is for humans. Equine FACS~\cite{wathan2015equifacs} annotators must be trained and certified, and spend upwards of 30 minutes to annotate a single minute of video~\cite{rashid2018should}. Second, with pain scales including entries like `interactive behavior' there is not always a mapping between pain attribute and obvious visual behavior~\cite{gleerup2016recognition} making detailed annotation of video an ambiguous task. Third, horses behave differently in the presence of humans~\cite{coles2016no, torcivia2021equine}, which calls to question the applicability of datasets with observed horses to more natural settings when the horse is alone. Relatedly, a vision based monitoring system for horses would be more pragmatic if it could operate off unobtrusive surveillance cameras, rather than require horses to be close to the camera, with the face clearly visible, as is true for human datasets.

In 2018-2019, a dataset of horses with induced orthopaedic pain was collected at Swedish University of Agricultural Sciences~\cite{ask2020identification}. It includes multi-view surveillance video footage of unobserved horses with pain labels from periodic pain assessments by expert veterinary researchers. Since the dataset contains few subjects (8), and lacks temporally detailed annotation, a fully and strongly supervised network is likely to overfit to the training data. At the same time, the network predictions would not be interpretable, and may use superfluous but correlated information, such as the lighting to determine the pain state of the horse.

On the other hand, self-supervised methods have been shown to disentangle semantically and visually meaningful properties from image data without the use of labels. Examples include disentangling the 3D normals, albedo, lighting, and alpha matte for faces~\cite{shu2017cvpr}, and disentangling pose and identity for facial recognition~\cite{tran2017disentangled}.

Our \textbf{key idea} is to use self-supervision to disentangle the visual properties a pain classification system should focus on, and then use the disentangled representation to identify pain. In this manner, we can reduce the likelihood of the model learning extraneous information to determine pain, and prevent overfitting to the training data (Figure~\ref{fig_egp_key_idea}). Additionally we use the observation that a painful horse may also exhibit non-painful behavior to formulate a novel loss that makes best use of the sparse pain annotation. 

We use a two step process for pain classification. In the first stage, we train a view synthesis model that, given an image of a horse from one angle learns to synthesize the scene from a different viewpoint~\cite{rhodin2018unsupervised}. We use an encoder-decoder architecture, and disentangle the horse pose, appearance, and background in the process. In the second stage, we use the learned pose representation to classify video segments as painful. As we lack detailed temporal annotation for pain, we use weak supervision to train the pain classification module, and propose a modified multiple instance learning loss towards this end. Our system is able to learn a viewpoint aware latent pose representation, and determine the pain label of video segments with 60\% accuracy. In comparison, human performance on a similar equine orthopaedic dataset is at 51\% accuracy~\cite{broome2021sharing}.

We present pain classification results of our model alongside ablation studies comparing the contribution of each module. In addition, we analyze the features of pain detected by our system, and note their correspondence with current veterinary knowledge on equine pain behavior.

Our contributions are:
\begin{itemize}
    \item Creating a disentangled horse pose representation from multi-view surveillance video footage of horses in box stalls using self-supervised novel view synthesis.
    \item Presenting a method for video segment level pain classification from the learned disentangled horse pose representation that is trained using weak pain labels and a novel multiple instance learning (MIL) loss.
    \item Extensive experiments including analysis of our automatically detected pain video segments in terms of cues used for pain diagnosis in veterinary practice.
\end{itemize}

\section{Related work}

Our work is closely related to novel view synthesis: given a view of a scene, the task is to generate images of the scene from new viewpoints. This is challenging, as it requires reasoning about the 3D structure and semantics of the scene from the input image. Rhodin et al.~\cite{rhodin2018unsupervised} train an encoder-decoder architecture on the novel view synthesis task using synchronized multi-view data to create a latent representation which disentangles human pose and appearance. Our work uses the same approach to learn a disentangled horse pose representation. However, while their method uses the learned pose representation for the downstream and strongly related task of 3D and 2D body keypoint estimation, our work uses the latent representation to classify animal pain in a weakly supervised setting.

Others achieve novel view synthesis assisted by either noisy and incomplete~\cite{choi2019extreme,sitzmann2019scene,novotny2019perspectivenet,hedman2018deep}, or ground truth depth maps, in addition to images during training~\cite{li2018megadepth,tulsiani2018factoring,shin20193d,niklaus20193d}.

Similar to our work, generative models have been used with emphasis on learning a 3D aware latent representation. Of note are deep voxels~\cite{sitzmann2019deepvoxels}, HoloGAN~\cite{nguyen2019hologan}, and  Synsin~\cite{wiles2020synsin}. Although we share the aim
to create a 3D aware latent representation, these methods emphasize the generation of accurate and realistic synthetic views, while our work focuses on using the latent representation for the downstream task of pain classification. While we do make use of multi-view data, the different viewpoints are few -- 4 -- and are separated by a wide baseline, unlike the above mentioned novel view synthesis works.

Generative models with disentangled latent representations have been developed for a wide range of purposes such as to discover intrinsic object properties like normals, albedo, lighting, and alpha matte for faces~\cite{shu2017cvpr}, fine grained class attributes for object discovery and clustering~\cite{singh2019finegan}, and pose invariant identity features for face recognition~\cite{tran2017disentangled}, and have been a topic of extensive research~\cite{tenenbaum2000separating,chen2016infogan,higgins2016beta,denton2017unsupervised,hu2018disentangling}. Our work relies on a disentangled pose representation from multi-view data, and places emphasis on utilizing the learned representation for a downstream task. Self-supervised disentangled pose representations have been used for 2D and 3D keypoint recognition~\cite{rhodin2018unsupervised,rhodin2019neural,chen2019weakly,chen2019unsupervised,habibie2019in}, but no previous work has used them for the behavior related task of pain recognition, particularly in animals. 

There is a growing body of work on deep visual learning for animal behavior and body understanding. This includes work on animal body keypoint prediction~\cite{cao2019cross,mu2020learning}, facial keypoint prediction~\cite{yang2016human,rashid2017interspecies,khan2020animal}, and dense 3D pose estimation via transfer from human datasets~\cite{sanakoyeu2020transferring} and fitting a known 3D model to 2D image~\cite{zuffi20173d,zuffi2018lions}. Of note are~\cite{zuffi2019three} which uses a synthesized zebra dataset to train a network for predicting zebra pose, and~\cite{li2021hsmal} which develops a horse specific 3D shape basis and applies it to the downstream task of lameness detection.

Beyond animal keypoint and pose prediction, there is a growing body of research on detecting animal affective states from images and videos. Of great relevance is Broom{\'e} et al.'s~\cite{broome2019dynamics} work on horse pain recognition that uses a fully recurrent network for pain prediction on horse videos. The method uses strong supervision and a dataset with close up videos of horses with acute pain. Follow up work~\cite{broome2021sharing} uses a similar network architecture and domain transfer from between acute and orthopaedic pain for horse pain classification. Sheep~\cite{lu2017estimating}, donkey~\cite{hummel2020automatic}, and mouse~\cite{tuttle2018deep} pain have also been explored with promising results. However, previous methods use either facial data, strong supervision, or additional information such as keypoints, segmentation masks, or facial movement annotation to learn the pain models. In contrast, we use weak supervision, with no additional annotation or data, and video data with the full body of the horse visible rather than just the face.

\section{Approach}
The equine pain dataset comprises time aligned videos of horses from multiple views, and pain labels from periodic observations. We use a two-step approach for pain classification. In the first stage, we train an encoder-decoder architecture for novel view synthesis, and learn an appearance invariant and viewpoint covariant horse pose latent representation in the process. In the second stage, we train a pain classification head using the trained pose aware latent representation as input to diagnose pain in video sequences. Since the dataset does not have detailed temporal annotation of horse pain expression, we use a MIL approach for pain classification for video sequences. In the following sections, we first describe the dataset, followed by details of our view synthesis and pain classification methods.

\subsection{Dataset}
The Equine Orthopaedic Pain (EOP) Dataset~\cite{ask2020identification} comprises of 24-hour surveillance footage of 8 horses filmed over 16 days before and during joint pain induction. The experimental protocol was approved by the Swedish Ethics Committee in accordance with the Swedish legislation on animal experiments (diary number 5.8.18-09822/2018). As few horses as possible were included and a fully reversible lameness induction model was used. Hindlimb lameness was induced by injecting lipopolysaccharides (LPS) into the hock joint, leading to inflammatory pain and various degrees of joint swelling. To decrease such pain, the horse tries to unload the painful limb both at rest and during motion, and movement asymmetry can then be measured objectively. The horses were stalled individually in one of two identical box stalls, with four surveillance cameras in the corners of each stall capturing round the clock footage. Starting 1.5 hours after joint injection, horses were periodically removed from the box stall to measure movement asymmetry during trot. Measurements were discontinued once horse movement asymmetry returned to levels similar to before LPS injection. In addition, pain assessments using four different pain scales~\cite{dalla2014development, gleerup2016recognition, van2015monitoring, bussieres2008development} were performed by direct observation 20 minutes before and after each movement asymmetry measurement. 

\subsubsection{Data preprocessing}
We use video data from two hours of pre-induction baseline as our no-pain data. Pain level was determined by averaging the Composite Pain Scale (CPS) score~\cite{bussieres2008development} of three veterinary experts during each pain assessment session. The session with the highest average CPS score was selected as peak pain period. Two hours closest to the pain observation session were used as our pain data. Four (two pain, two no-pain) hours per horse, from four surveillance cameras for eight horses results in a total of 128 hours of film. We only use the time periods when no humans are present in the stall or the corridor outside the stall, reducing the likelihood of interactivity leading to changes in the horses’ behavior.

Videos were cut in to 2 minute long segments (details in Section~\ref{sec_implementation}), and all segments belonging to the two hour peak pain period were labeled as pain, and pre-induction period were labeled as no-pain. 

Note that we extrapolate pain labels from the pain observation session to the closest unobserved horse videos. These videos have not been viewed and annotated as containing pain behavior by experts, and the typical duration and frequency of horse pain expression is not known. Similarly, the no-pain videos were not manually verified as not containing any pain expression. As a result, our video level pain labels are likely noisy and contribute to the difficulty of our task.

\subsection{Multi-view synthesis}\label{section_base_net}
The multi-view synthesis network uses the same architecture and training methodology as original work by Rhodin et al~\cite{rhodin2018unsupervised}: a U-Net architecture~\cite{ronneberger2015u} learns a disentangled horse appearance and pose representation in its bottleneck layer. This is done by training the model to synthesize an input scene from a different viewpoint.

Specifically, given an input video frame $x_{i,t}$, from viewpoint $i$, at time $t$, the encoder, $f_E$, outputs are a latent representation of the horse pose, $p_{i,t}$, and of the horse appearance, $h_{i,t}$:
\begin{equation}
p_{i,t},\,h_{i,t}\,=\,f_E\,(x_{i,t}).
\end{equation}

During training, both $p_{i,t}$ and $h_{i,t}$ are manipulated before being input
to the decoder, $f_D$. The pose representation is multiplied by the relative camera rotation, $\mathbf{R}$, of viewpoint $i$ and $j$, so that the pose representation input to $f_D$ is in the same space as pose representations for viewpoint $j$:
\begin{equation}
p_{i\rightarrow j,t}\,=\,\mathbf{R}_{v\rightarrow j}\,p_{i,t}.
\end{equation}

The appearance representation is swapped by the appearance representation of an input frame of the same horse from the same viewpoint, but from a different time, $t'$, and hence likely with a different pose. This appearance representation swap encourages the network to disentangle pose and appearance. In addition, a background image for each viewpoint, $b_{i}$, is input to the decoder so that the network does not focus on learning background information for synthesis and instead focuses on the horse: 

\begin{equation}
x_{i\rightarrow j,t}'\,=\,f_D\,(p_{i\rightarrow j,t},\,h_{i,t'},\,b_{i}).
\end{equation}

As in~\cite{rhodin2018unsupervised}, the synthesized image, $x_{i\rightarrow j,t}'$, is supervised by both a pixel-level mean squared error (MSE) loss compared with the ground truth image, $x_{j,t}$, as well as a perceptual loss on ImageNet pretrained ResNet18~\cite{he2016deep} penultimate layer's features.
\begin{equation}
\begin{split}
L_{MVS} = ||x_{i\rightarrow j,t}'\,-\,x_{j,t}||^2\,+\, \\
\alpha\, ||\theta_{RN}(x_{i\rightarrow j,t}')\, - \,\theta_{RN}(x_{j,t})||^2,
\end{split}
\end{equation}

\noindent where $\alpha$ is a loss weighting parameter, and $\theta_{RN}(x)$ outputs ResNet18's penultimate feature representation for input image $x$. The multi-view synthesis loss, $L_{MVS}$, is averaged across all instances in a batch before backward propagation.

During testing, the synthesized image is generated without swapping of the appearance representation.

Similar to~\cite{rhodin2018unsupervised}, we detect and crop the horse in each frame to factor out scale and global position. $\mathbf{R}_{i\rightarrow j}$ is calculated with respect to the crop center instead of the image center. The crop is sheared so that it appears taken from a virtual camera pointing in the crop direction. In more detail, the crop is transformed by the homography induced by rotating the camera so that the ray through its origin aligns with the crop center.



\subsubsection{Refining multi-view synthesis for horses}\label{section_backbone_mod}
Unlike the Human 3.6 dataset~\cite{h36m_pami} used in~\cite{rhodin2018unsupervised}, which features actors that move around constantly, the EOP dataset features the horse standing or grazing in similar pose for long periods of time. As a result, randomly selecting a frame for the appearance feature swap on EOP can lead to suboptimal appearance disentanglement. Therefore, we pre-select time sequences with a variety of horse poses for training, based on the optical flow magnitude.

The EOP dataset was collected over multiple months, during which the cameras, although fixed as firmly as possible, were nudged by chewing from curious horses. As a result, background images calculated as the median image over the entire dataset were blurry. We therefore extracted separate higher quality background images per month.




\subsection{Classifying pain}\label{section_pain_head}
\begin{figure*}
    \centering
    \includegraphics[width=0.95\textwidth]{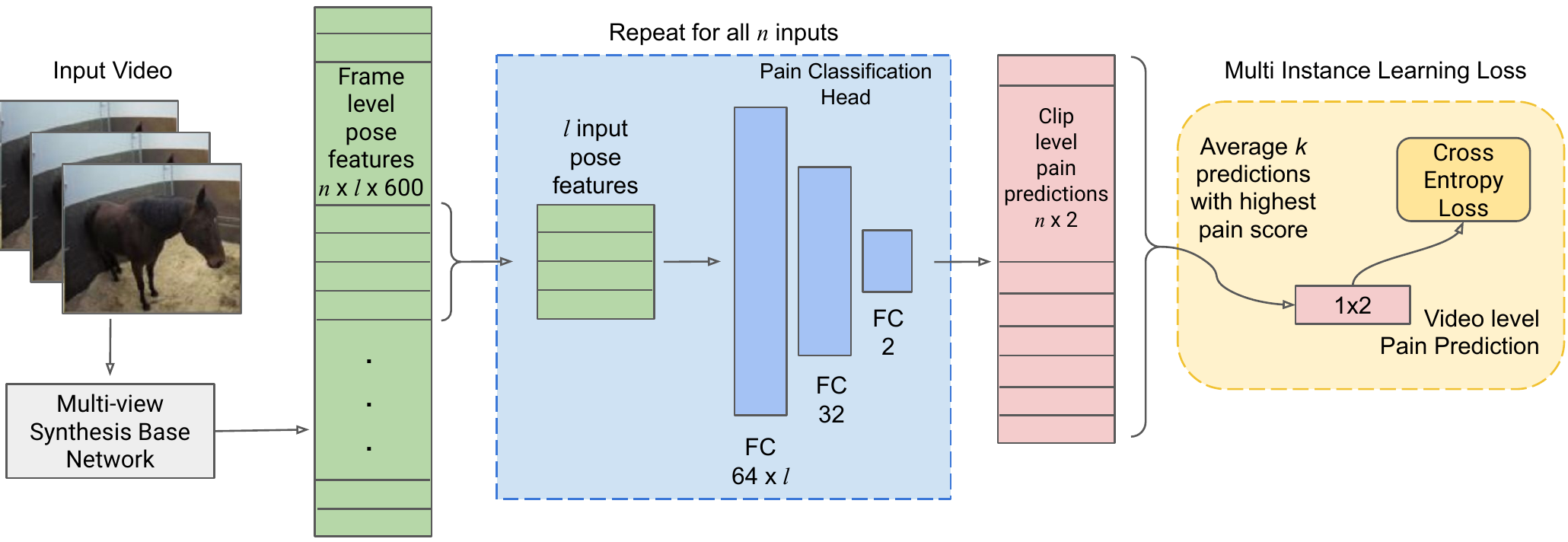}
    \caption{\textbf{Pain Classification Model.} Pose representations from the trained multi-view synthesis model are extracted for each frame of an input video, and collated into $l$ length clips. The pain classification head, comprising three fully-connected (FC) layers,  predicts pain for each clip. Clip level predictions are collated and supervised by the proposed MIL loss.}
    \label{fig_egp_overview}
\end{figure*}

The self-supervised base network (Section~\ref{section_base_net}) provides us with a means to disentangle the horse pose from its background, and appearance from a given input image. This representation is now used to train a pain classification head.

We treat video pain classification as multiple instance learning (MIL) problem. Every video comprises time segments that are independently classified as pain or no pain. These time segment level predictions are collated to obtain a video level pain prediction.

We classify pain classification with both frame and, following insight from earlier works~\cite{broome2019dynamics,rashid2020equine} showing pain classification from instantaneous observations to be unreliable, clip level inputs. For the latter, we concatenate per-frame latent representations into a clip level volume used as the atomic unit for pain classification during both training and testing.


The network architecture, shown in Figure~\ref{fig_egp_overview}, comprises two hidden linear layers with ReLU and dropout, followed by a classification layer with two dimensional output for pain vs no-pain. Frame level predictions from the first linear layer are concatenated in to a $64*l$ vector, where $l$ is the number of frames in each input segment, that is then forwarded through the network. 

More specifically, each video sequence $s$, comprises $n$ time segments indexed by $t$. The pain head, $\theta$, provides a two-dimensional output that is softmaxed to obtain the pain, and no pain confidence values for time segment $t$, where $\mathbf{p}$ represents the set of pose representations of all frames in $t$:
\begin{equation}
y_{i,t}^{NP}\,,\,y_{i,t}^P\,=\, \text{softmax}\,(\theta\,(\mathbf{p}_{i,t})).
\end{equation}

The $k$ time segments with the highest \emph{pain} predictions are averaged to obtain the video level pain prediction: 
\begin{equation}
y_{i,s}^P\,=\,\frac{1}{k}\,\sum_{t \in S}\,y_{i,t}^P~,~~y_{i,s}^{NP}\,=\,\frac{1}{k}\,\sum_{t \in S}\,y_{i,t}^{NP},
\end{equation}

\noindent where $S$ is the set of $k$ time segments' indices with the highest pain prediction:
\begin{equation}
S = \{j\,|\,y_{i,j}^P \in \max_{K}\,\{y_{i,1}^P,y_{i,2}^P,\hdots,y_{i,n}^P\}\}.
\end{equation}

The video level pain predictions are then supervised with a cross-entropy loss.

Parameter $k$ is set to $\lfloor\frac{n}{d}\rfloor$, where $d$ is randomly selected from $\{1,2,4,8\}$ at every training iteration, and set to 8 during testing. $d$ correlates with the proportion of a video that is predicted as an action class. As we do not know the proportion of time a horse in pain will express pain in a video, varying this parameter randomly is likely to provide the most robust performance, as shown in previous work~\cite{rashid2020action}.

Our key insight in designing the loss is that a horse in pain need not express pain constantly, and the absence of its pain behavior can be rightly classified as no-pain. By collating only the predictions for the top $k$ \textit{pain} segments to obtain the video level prediction, we do not penalize the network for classifying no-pain time segments within a pain video. We hence require only that the pain predictions have high confidence in pain videos, and that no time segments have high pain confidence in no-pain videos. Our loss is different from the MIL loss used in literature (e.g.~\cite{wang2017untrimmednets,paul2018wtalc,nguyen2018weakly}, which would have averaged the highest $k$ predictions for both pain and no-pain class independently. Section~\ref{section_egp_mil_exp} compares these loss formulations. 

\section{Experiments}
\subsection{Implementation details}\label{sec_implementation}
Optical flow was calculated using Farneb{\"a}ck's method~\cite{farneback2003two} on video frames extracted at 10 frames per second. Time segments that had optical flow magnitude in the top 1\%, 143559 frames, were used to train the multi-view synthesis module. We use leave-one-out, subject exclusive training. Multi-view synthesis (backbone) models were trained for 50 epochs at 0.001 learning rate using Adam optimizer. The perceptual loss is weighted 2 times higher than the MSE loss during training. The same U-Net based architecture as in~\cite{rhodin2018unsupervised} is used. The 600 dimensional pose representation is reshaped to $200\times3$ for multiplication with the rotation transformation $\mathbf{R}$.

The pain classification dataset comprises video segments, $s$, with the maximum length of 2 minutes. MaskRCNN~\cite{he2017mask} was used to detect, crop and center the horse in each frame. Missing MaskRCNN detections can reduce the video segment length, but no instances less than 10 seconds in length are included. Note that all data with valid horse detections were included in the pain classification dataset, and not just the high motion frames.

Pain is predicted for short clips, of length $l$, that are collated for video level pain prediction. We show results when using clips of length 1 frame (frame based), and with clips of length five seconds at 2 fps. The five second clip length is set following past research~\cite{broome2019dynamics}; additionally, \cite{rashid2020equine} suggests it to be the duration of time a horse pain expression lasts.

The backbone network is frozen when training the pain classification head, which is trained for 10 epochs at 0.001 learning rate. Leave-one-out training is again used, excluding the same test subject as for the backbone network. In addition, pain classification performance on a validation set is calculated after each epoch, and the model at the epoch with the highest performance is used for testing. Data from the non-test subject with the most balanced pain/no-pain data distribution is used as the validation data. Further details are included in the supplementary.

\subsection{Disentangled representation learning}
\begin{figure*}[t]
    \centering
    \includegraphics[width=\textwidth]{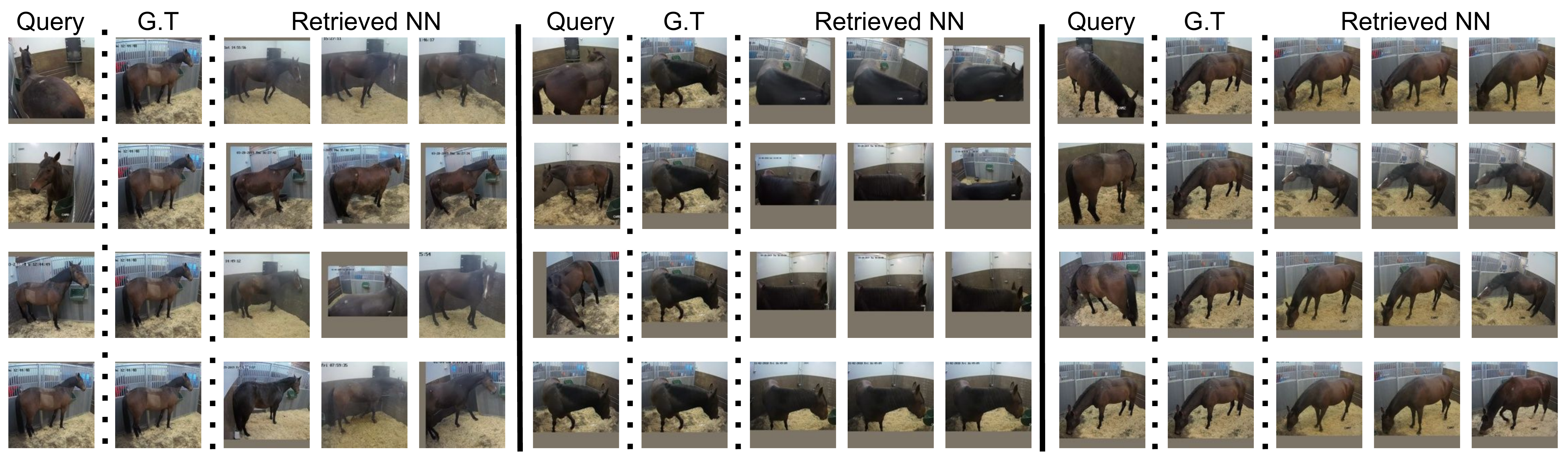}
    \caption{Nearest neighbor retrieval on latent pose representation. The pose representation of the query image is rotated before nearest neighbor retrieval. The nearest neighbors match the pose in the actual ground truth image from the rotated view.}
    \label{fig_horse_nn}
\end{figure*}
\begin{figure}[t]
    \centering
    \includegraphics[width=\linewidth]{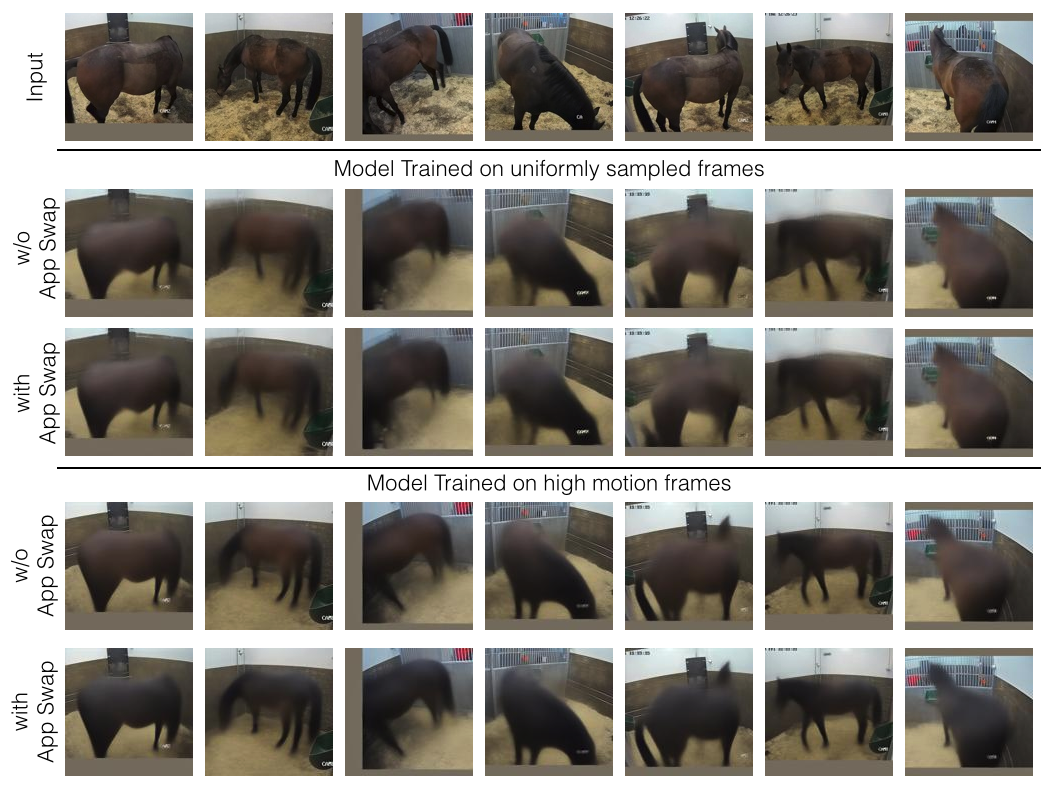}
    \caption{Appearance (App) swapping on different models. Each column shows the decoder output for the corresponding input image from the first row. In the third and fifth rows, the appearance representations are swapped with the appearance representation from a training horse with a black coat.}
    \label{fig_app_swap}
\end{figure}
\paragraph{Disentangled pose representation.}
We explore the quality of the latent representation qualitatively. The ideal pose representation would be able to cluster the same horse pose regardless of viewpoint. In addition, this representation would be disentangled from horse appearance. 

Given the pose representation of a test input image at time $t$ from viewpoint $i$, $p_{i,t}$, we find its top 3 nearest neighbors in the training data after rotation to viewpoint $j$. That is, we find the nearest neighbors of $\mathbf{R}_{i\rightarrow j}\,p_{i,t}$. Some qualitative results are shown in Figure~\ref{fig_horse_nn}, where the second columns show the actual image from viewpoint $j$. 

The top 3 neighbors are consistent with the expected ground truth, showing
that the network has learned a pose representation that is viewpoint covariant. One exception is the second nearest neighbor in the third row, left set, that is quite different from the ground truth image. The backgrounds of the retrieved images are often different from the query background, for example in the middle set, showing background disentanglement. Moreover, the retrieved horses may be physically different from the query horse, showing appearance disentanglement: a black horse is retrieved in the fourth row, left set, and a horse with a white blaze is retrieved in the second row, right set. Interestingly, when the horse head and neck is occluded in the second row, right set, the nearest neighbors suggest that the model hallucinates a reasonable -- though not entirely accurate -- neck and head position.

\paragraph{Disentangled identity representation.}
In Figure~\ref{fig_app_swap}, we show results of swapping the appearance representation. As explained in Section~\ref{section_base_net}, the decoder uses an appearance and pose representation to reconstruct an image. We compare reconstructed test images with and without swapping the appearance representations with the appearance representation of a training horse with a black coat. Good disentanglement would show a horse with the same pose as the input image, but with a black rather than a brown coat.

The model trained with uniformly sampled video frames is not able to disentangle appearance and pose and reconstructs horses with more or less the same color, both with and without appearance swapping. High motion sampling during training increases the chance that the swapped appearance frame features a horse in a different pose than the input frame. As seen by the darker coats in the last row, it hence leads to better appearance disentanglement. It also leads the network to learn a variety of poses, as can be seen by the crisper reconstructions around the head and legs, in the fifth and sixth columns. Lastly, the background is crisper in the last two rows. This is due to our use of crisper background images derived from the same month as the input that results in better background disentanglement.

\subsection{Pain classification}
\begin{table}[t]
\centering
\resizebox{0.9\linewidth}{!}{
    	\begin{tabular}{|c|c|c|c|c|}
    	\hline
		& \multicolumn{2}{c|}{True Performance} & \multicolumn{2}{c|}{Oracle Performance}\\
		\hline
		& F1 Score & Accuracy & F1 Score & Accuracy\\
		\hline
		Ours-Frame & \textbf{58.5$\pm$7.8} & \textbf{60.9$\pm$5.7} & 60.8$\pm$6.4 & 62.3$\pm$5.4\\
        Ours-Clip & 55.9$\pm$5.1 & 57.8$\pm$4.4 & \textbf{65.1$\pm$6.7} & \textbf{65.6$\pm$6.5}\\
        Ours-Clip-HaS & 56.5$\pm$5.0 & 58.6$\pm$4.3 & 63.6$\pm$6.2 & 64.6$\pm$5.8\\
        Scratch & 54.5$\pm$9.1 & 57.3$\pm$6.5 & 61.7$\pm$8.1 & 63.2$\pm$7.7\\
        Broom{\'e} `19~\cite{broome2019dynamics} & 53.0$\pm$8.1 & 55.2$\pm$7.0 & 60.0$\pm$8.3 & 59.4$\pm$8.3\\
        \hline
	    \end{tabular}}
\caption{Comparison of frame and clip based pain heads against a models trained from scratch with early stopping using a hold out dataset (True Performance) and best case (Oracle).}
\label{table_mil}       
\end{table}

\begin{table}[t]
\begin{minipage}{\linewidth}
\centering
\resizebox{.45\textwidth}{!}{
    \centering
    	\begin{tabular}{|c|c|c|}
    	\hline
        & F1 Score & Accuracy\\
        \hline
        Ours-Frame & \textbf{58.5$\pm$7.8} & \textbf{60.9$\pm$5.7}\\
        NoApp & 57.1$\pm$7.9 & 59.7$\pm$6.0\\
        NoBG & 53.2$\pm$8.3 & 55.9$\pm$7.4 \\
        NoApp+NoBG & 55.6$\pm$5.2 & 57.8$\pm$5.7\\
        NoMod & 56.6$\pm$7.4 & 58.4$\pm$5.5\\
		\hline
	    \end{tabular}}
\resizebox{.45\textwidth}{!}{
    \centering
    	\begin{tabular}{|c|c|c|}
    	\hline
        & F1 Score & Accuracy\\
        \hline
        Ours-MIL & \textbf{58.5$\pm$7.8} & \textbf{60.9$\pm$5.7}\\
        CE-Clip & 52.2$\pm$10.2 & 57.0$\pm$6.4 \\
        CE-Frame & 49.1$\pm$10.9 & 55.2$\pm$5.9\\
        MIL-OG & 47.7$\pm$12.7 & 55.0$\pm$8.2\\
		\hline
	    \end{tabular}}
\end{minipage}
\caption{\emph{Left}: Pain classification performance with backbones of varying pose disentanglement. \emph{Right}: Comparison of cross-entropy loss and multi instance learning loss variations.}
\label{table_ablations}       
\end{table}

We present F1 score and accuracy for pain classification results, taking the unweighted mean of the F1 score across both classes. Both metrics are averaged across all training folds, and presented here alongside the standard deviation. 

In Table~\ref{table_mil} we compare variants of our pain classification head. The `True Performance' column shows the performance of the model selected by early stopping based on performance on a holdout validation set. `Oracle Performance' shows results if the early stopping criteria aligned with the epoch with the best testing performance and shows the upper limit performance. `Ours-Frame' uses frame level inputs ($l=1$), `Ours-Clip' uses 5 second clips. The `Scratch' model has the same architecture as the encoder part of the base network and the pain head and is trained on frame level inputs, supervised by our MIL loss, and is trained from scratch. 



Both our frame and clip based models have better true performance than the scratch model; the frame based model shows 4 percentage points (pp.) higher F1 score. Additionally, the oracle performance is either comparable or better than the scratch model, even though the latter learns more parameters specific to the pain classification task. At the same time, our models' use of a disentangled pose representation ensures that pose rather than extraneous information is used to deduce the pain state. These results indicate that using a disentangled pose representation is useful for a dataset such as ours with limited training subjects.

All models suffer from some degree of overfitting, as can be seen from the difference between the true and oracle results. The clip based and scratch models exhibit a high degree of overfitting: with oracle F1 score at $\sim10$ and $\sim7$ pp. higher than the true performance. This can be expected since these models learns more parameters, and points to the need for a light weight pain classification head. 

However, the use of temporal information through clip level inputs results in a much higher oracle performance compared to frame level inputs (5 pp. higher F1). We therefore add more regularization by using random adversarial erasing using Hide-and-Seek ~\cite{kumar2017hide} (`Ours-Clip-HaS') resulting in 1 pp higher accuracy. Altogether, clip based pain prediction is most promising, but would require more regularization to compete with the simple frame based results.

Finally, `Broom{\'e} `19' shows results of the convolution LSTM model proposed in ~\cite{broome2019dynamics} on EOP dataset. As we do not use optical flow frames in other experiments, we use the one-stream (RGB) variant of their model. Following the original work we evaluate and train on 10 frame clips at 2 FPS, supervised with binary cross entropy loss. The method achieves performance slightly worse than the `Scratch' model, and shows similar overfitting behavior.

\paragraph{Backbone Ablation study.}\label{section_backbone_ablation_exp}
In Table~\ref{table_ablations} (left) we show the results on an ablation study on our backbone model. Each model variant has the same pain detection head as `Ours-Frame', but uses different backbone trainings. `NoBG' means that the backbone was trained without providing a background frame to the decoder and has low background disentanglement. `No-App' means that the appearance feature swapping was not used during training, and appearance disentanglement is low. Finally, `No-Mod' does not use our modifications mentioned in Section~\ref{section_backbone_mod} and shows the importance of motion based sampling, and a better background, but is otherwise identical to `Ours-Frame'. 

Each component of the backbone is important for final pain results, and shows the importance of a well disentangled pose representation for pain classification. As shown in Figure~\ref{fig_app_swap}, `No-Mod' has little appearance disentanglement and shows similar results to `No-App', but is slightly worse probably because it also uses worse background images. Background disentanglement is most important for our performance, with `NoBG' achieving ~5 pp. poorer performance than our full model. Interestingly, when both background image and appearance swapping are removed (`NoBG+NoApp'), the model does better than when just background image is removed (`NoBG'). This may be because appearance swapping prevents the model from encoding any background knowledge in the appearance latent features, making it harder for the model to disentangle background on its own.

\paragraph{Weakly supervised learning.}\label{section_egp_mil_exp}
\begin{figure}[t]
    \centering
    \includegraphics[width=0.9\linewidth]{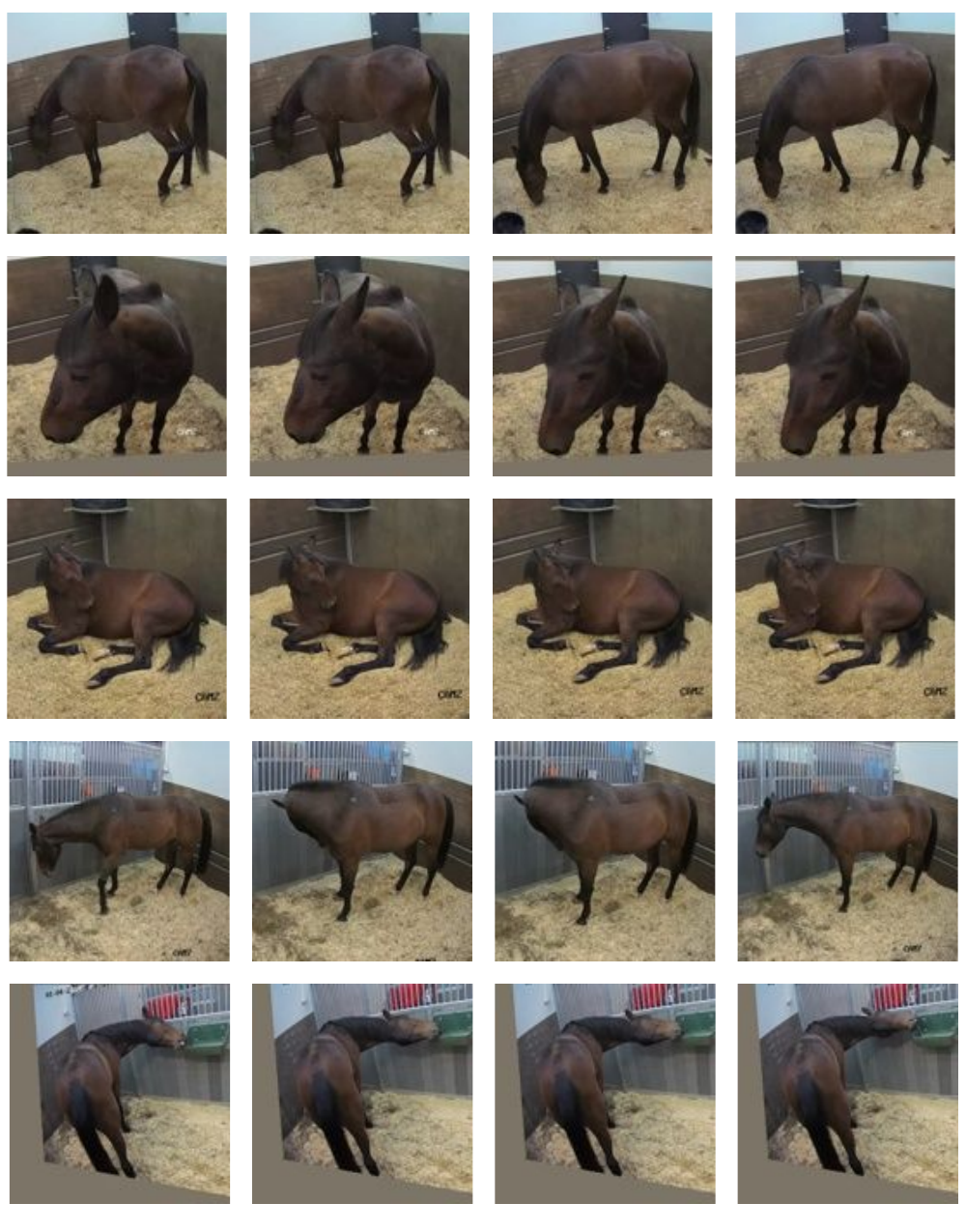}
    \caption{Video segments correctly detected as painful by our model. The detected segments display signs of pain such as avoiding weight bearing (1st row), backwards ears (2nd row), lying down (3rd row), looking at the painful leg (4th row), and stretching (last row).}
    \label{figure_pain_segs}
\end{figure}

In Table ~\ref{table_ablations} (right), we evaluate the importance of multi-instance learning. As discussed in Section~\ref{section_pain_head}, our version of MIL loss (Ours-MIL), averages the pain and no-pain predictions of the top $k$ time segments (or frames) with the highest \emph{pain} prediction. We contrast this against the original MIL loss (MIL-OG), which averages the top pain and no-pain predictions separately. Lastly, we compare against a simple cross-entropy loss (CE), where each frame or clip is separately supervised during training. The test results are still obtained by averaging the top $k$ clip level predictions to keep results comparable.

Firstly, by comparing Ours-MIL against CE we see that using a MIL setting is essential, and that pain (or no-pain) behavior is not present in each frame from a pain (or no-pain) video. In fact, the results are lower than random for a CE-trained model (49.1\% F1 score). The use of dynamic information with clip based inputs leads to improved performance, although the overall performance is lower than for weakly supervised training. Secondly, compared to MIL-OG, we see that Ours-MIL is necessary to learn a reasonable model for pain. In fact, MIL-OG leads to a worse model of pain than random guessing, at $47.7\%$ F1 score. This bolsters our underlying reasoning that clips with no-pain features may exist in pain videos and should not be penalized during training in order to develop a good pain model.

\subsection{Attributes of pain}
Figure~\ref{figure_pain_segs} shows example clips that our model classifies as painful. The clips contain classic signs of pain such as `lowered ears'~\cite{gleerup2015equine} (second and fourth row), a lifted left hind limb (first row), corresponding to `non-weight bearing'~\cite{bussieres2008development}, `lying down' (third row), `looking at flank' (third and fourth row) ~\cite{pritchett2003identification}, and an example of gross pain behavior, `stretching' (last row)~\cite{gleerup2016recognition}. These results show a good correspondence between the visual attributes our model focuses on, and pain scales used by veterinary experts. While we only expected body behavior to be picked up by the pain model, interestingly, subtle facial behavior, specifically, ear movements, are also picked up and learned. More results are shown in the supplementary.

\section{Discussion}
Equine pain detection is an intrinsically challenging problem. Recent work~\cite{broome2021sharing} shows that equine orthopaedic pain is particularly difficult to detect, as it has high signal to noise ratio and therefore requires transfer learning from cleaner acute pain data for reasonable performance (49.5\% F1 before transfer learning, and 58.2\% after). EOP dataset presents additional challenges as pain labels are sparse and noisy. Despite these challenges, our method can achieve 60\% accuracy which is better than human expert performance on equine orthopaedic pain (average of 51.3\% accuracy across three pain levels~\cite{broome2021sharing}), and on par with human expert performance on acute pain~\cite{broome2019dynamics}. 

Our method is scalable and pragmatic as we use unobtrusive surveillance footage, and sparse pain labels. With few training subjects, and noisy labels, we ensure that pain is learned from horse body language alone by use of a self-supervised generative model to disentangle the horse pose from appearance and background. The resulting pose representation is used to learn a pain prediction model, weakly supervised with a novel pain specific multiple instance learning loss. We qualitatively analyze our model's disentangled pose and appearance features, and show quantitative and qualitative results on pain classification.

Future work should include means to exclude pain predictions on videos without a clear enough view of the horse. As we do not know the typical frequency and duration of horse pain expression, the video length used in this work may be sub-optimal. Research on the optimal duration of a video segment to guarantee the observation of pain behavior, and further regularization of the pain classification head are promising directions of future improvement.
{\small
\bibliographystyle{ieee_fullname}
\bibliography{egbib}

\begin{thebibliography}{10}\itemsep=-1pt

\bibitem{andersen2018can}
Pia~H Andersen, KB Gleerup, J Wathan, B Coles, H Kjellstr{\"o}m, S Broom{\'e},
  YJ Lee, M Rashid, C Sonder, E Rosenberg, and D Forster.
\newblock Can a machine learn to see horse pain? an interdisciplinary approach
  towards automated decoding of facial expressions of pain in the horse.
\newblock In {\em Measuring Behavior}, 2018.

\bibitem{ask2020identification}
Katrina Ask, Marie Rhodin, Lena-Mari Tamminen, Elin Hernlund, and Pia
  Haubro~Andersen.
\newblock Identification of body behaviors and facial expressions associated
  with induced orthopedic pain in four equine pain scales.
\newblock {\em Animals}, 10(11), 2020.

\bibitem{opencv_library}
G. Bradski.
\newblock {The OpenCV Library}.
\newblock {\em Dr. Dobb's Journal of Software Tools}, 2000.

\bibitem{broome2021sharing}
Sofia Broom{\'e}, Katrina Ask, Maheen Rashid, Pia~Haubro Andersen, and Hedvig
  Kjellstr{\"o}m.
\newblock Sharing pain: Using domain transfer between pain types for
  recognition of sparse pain expressions in horses.
\newblock {\em arXiv preprint arXiv:2105.10313}, 2021.

\bibitem{broome2019dynamics}
Sofia Broom{\'e}, Karina~Bech Gleerup, Pia~Haubro Andersen, and Hedvig
  Kjellstr{\"o}m.
\newblock Dynamics are important for the recognition of equine pain in video.
\newblock In {\em CVPR}, 2019.

\bibitem{bussieres2008development}
G Bussieres, C Jacques, O Lainay, G Beauchamp, Agn{\`e}s Leblond, J-L
  Cador{\'e}, L-M Desmaizi{\`e}res, SG Cuvelliez, and E Troncy.
\newblock Development of a composite orthopaedic pain scale in horses.
\newblock {\em Research in veterinary science}, 85(2), 2008.

\bibitem{cao2019cross}
Jinkun Cao, Hongyang Tang, Hao-Shu Fang, Xiaoyong Shen, Cewu Lu, and Yu-Wing
  Tai.
\newblock Cross-domain adaptation for animal pose estimation.
\newblock In {\em ICCV}, 2019.

\bibitem{chen2019unsupervised}
Ching-Hang Chen, Ambrish Tyagi, Amit Agrawal, Dylan Drover, Rohith MV, Stefan
  Stojanov, and James~M. Rehg.
\newblock Unsupervised {3D} pose estimation with geometric self-supervision.
\newblock In {\em CVPR}, 2019.

\bibitem{chen2016infogan}
Xi Chen, Yan Duan, Rein Houthooft, John Schulman, Ilya Sutskever, and Pieter
  Abbeel.
\newblock Infogan: Interpretable representation learning by information
  maximizing generative adversarial nets.
\newblock {\em Advances in neural information processing systems}, 29, 2016.

\bibitem{chen2019weakly}
Xipeng Chen, Kwan-Yee Lin, Wentao Liu, Chen Qian, and Liang Lin.
\newblock Weakly-supervised discovery of geometry-aware representation for 3d
  human pose estimation.
\newblock In {\em CVPR}, 2019.

\bibitem{choi2019extreme}
Inchang Choi, Orazio Gallo, Alejandro Troccoli, Min~H Kim, and Jan Kautz.
\newblock Extreme view synthesis.
\newblock In {\em ICCV}, 2019.

\bibitem{coles2016no}
Britt~Alice Coles.
\newblock No pain, more gain? evaluating pain alleviation post equine
  orthopedic surgery using subjective and objective measurements.
\newblock {\em Swedish University of Agricultural Sciences, Masters Thesis},
  2016.

\bibitem{dalla2014development}
Emanuela Dalla~Costa, Michela Minero, Dirk Lebelt, Diana Stucke, Elisabetta
  Canali, and Matthew~C Leach.
\newblock Development of the {H}orse {G}rimace {S}cale ({HGS}) as a pain
  assessment tool in horses undergoing routine castration.
\newblock {\em PLoS one}, 9(3), 2014.

\bibitem{denton2017unsupervised}
Emily~L Denton et~al.
\newblock Unsupervised learning of disentangled representations from video.
\newblock In {\em Advances in neural information processing systems}, 2017.

\bibitem{ekman2002facial}
Paul Ekman.
\newblock Facial action coding system ({FACS}).
\newblock {\em A human face}, 2002.

\bibitem{farneback2003two}
Gunnar Farneb{\"a}ck.
\newblock Two-frame motion estimation based on polynomial expansion.
\newblock In {\em Scandinavian conference on Image analysis}. Springer, 2003.

\bibitem{gleerup2016recognition}
KB Gleerup and Casper Lindegaard.
\newblock Recognition and quantification of pain in horses: A tutorial review.
\newblock {\em Equine Veterinary Education}, 28(1), 2016.

\bibitem{gleerup2015equine}
Karina~B Gleerup, Bj{\"o}rn Forkman, Casper Lindegaard, and Pia~H Andersen.
\newblock An equine pain face.
\newblock {\em Veterinary anaesthesia and analgesia}, 42(1), 2015.

\bibitem{graubner2011clinical}
Claudia Graubner, Vinzenz Gerber, Marcus Doherr, and Claudia Spadavecchia.
\newblock Clinical application and reliability of a post abdominal surgery pain
  assessment scale (paspas) in horses.
\newblock {\em The Veterinary Journal}, 188(2), 2011.

\bibitem{habibie2019in}
Ikhsanul Habibie, Weipeng Xu, Dushyant Mehta, Gerard Pons-Moll, and Christian
  Theobalt.
\newblock In the wild human pose estimation using explicit {2D} features and
  intermediate {3D} representations.
\newblock In {\em CVPR}, 2019.

\bibitem{he2017mask}
Kaiming He, Georgia Gkioxari, Piotr Doll{\'a}r, and Ross Girshick.
\newblock Mask rcnn.
\newblock In {\em ICCV}, 2017.

\bibitem{he2016deep}
Kaiming He, Xiangyu Zhang, Shaoqing Ren, and Jian Sun.
\newblock Deep residual learning for image recognition.
\newblock In {\em CVPR}, 2016.

\bibitem{hedman2018deep}
Peter Hedman, Julien Philip, True Price, Jan-Michael Frahm, George Drettakis,
  and Gabriel Brostow.
\newblock Deep blending for free-viewpoint image-based rendering.
\newblock {\em ACM Transactions on Graphics (TOG)}, 37(6), 2018.

\bibitem{higgins2016beta}
Irina Higgins, Loic Matthey, Arka Pal, Christopher Burgess, Xavier Glorot,
  Matthew Botvinick, Shakir Mohamed, and Alexander Lerchner.
\newblock beta-vae: Learning basic visual concepts with a constrained
  variational framework.
\newblock In {\em ICLR}, 2016.

\bibitem{hu2018disentangling}
Qiyang Hu, Attila Szab{\'o}, Tiziano Portenier, Paolo Favaro, and Matthias
  Zwicker.
\newblock Disentangling factors of variation by mixing them.
\newblock In {\em CVPR}, 2018.

\bibitem{hummel2020automatic}
Hilde~I Hummel, Francisca Pessanha, Albert~Ali Salah, Thijs~JPAM van Loon, and
  Remco~C Veltkamp.
\newblock Automatic pain detection on horse and donkey faces.
\newblock In {\em FG}, 2020.

\bibitem{h36m_pami}
Catalin Ionescu, Dragos Papava, Vlad Olaru, and Cristian Sminchisescu.
\newblock Human3.6m: Large scale datasets and predictive methods for {3D} human
  sensing in natural environments.
\newblock {\em IEEE Transactions on Pattern Analysis and Machine Intelligence},
  36(7):1325--1339, 2014.

\bibitem{ireland2012comparison}
JL Ireland, PD Clegg, CM McGowan, SA McKane, KJ Chandler, and GL Pinchbeck.
\newblock Comparison of owner-reported health problems with veterinary
  assessment of geriatric horses in the united kingdom.
\newblock {\em Equine veterinary journal}, 44(1), 2012.

\bibitem{khan2020animal}
Muhammad~Haris Khan, John McDonagh, Salman Khan, Muhammad Shahabuddin, Aditya
  Arora, Fahad~Shahbaz Khan, Ling Shao, and Georgios Tzimiropoulos.
\newblock Animalweb: A large-scale hierarchical dataset of annotated animal
  faces.
\newblock In {\em CVPR}, 2020.

\bibitem{kumar2017hide}
Krishna Kumar~Singh and Yong Jae~Lee.
\newblock Hide-and-seek: Forcing a network to be meticulous for
  weakly-supervised object and action localization.
\newblock In {\em ICCV}, 2017.

\bibitem{li2021hsmal}
Ci Li, Nima Ghorbani, Sofia Broom{\'e}, Maheen Rashid, Michael~J Black, Elin
  Hernlund, Hedvig Kjellstr{\"o}m, and Silvia Zuffi.
\newblock {hSMAL}: Detailed horse shape and pose reconstruction for motion
  pattern recognition.
\newblock In {\em CV4Animals Workshop, CVPR}, 2021.

\bibitem{li2018megadepth}
Zhengqi Li and Noah Snavely.
\newblock Megadepth: Learning single-view depth prediction from internet
  photos.
\newblock In {\em CVPR}, 2018.

\bibitem{lu2017estimating}
Yiting Lu, Marwa Mahmoud, and Peter Robinson.
\newblock Estimating sheep pain level using facial action unit detection.
\newblock In {\em FG}. IEEE, 2017.

\bibitem{lucey2011painful}
Patrick Lucey, Jeffrey~F Cohn, Kenneth~M Prkachin, Patricia~E Solomon, and Iain
  Matthews.
\newblock Painful data: The {UNBC-McMaster} shoulder pain expression archive
  database.
\newblock In {\em FG}, 2011.

\bibitem{mu2020learning}
Jiteng Mu, Weichao Qiu, Gregory~D Hager, and Alan~L Yuille.
\newblock Learning from synthetic animals.
\newblock In {\em CVPR}, 2020.

\bibitem{nguyen2018weakly}
Phuc Nguyen, Ting Liu, Gautam Prasad, and Bohyung Han.
\newblock Weakly supervised action localization by sparse temporal pooling
  network.
\newblock In {\em CVPR}, 2018.

\bibitem{nguyen2019hologan}
Thu Nguyen-Phuoc, Chuan Li, Lucas Theis, Christian Richardt, and Yong-Liang
  Yang.
\newblock {HoloGAN}: Unsupervised learning of {3D} representations from natural
  images.
\newblock In {\em ICCV}, 2019.

\bibitem{niklaus20193d}
Simon Niklaus, Long Mai, Jimei Yang, and Feng Liu.
\newblock {3D} ken burns effect from a single image.
\newblock {\em ACM Transactions on Graphics (TOG)}, 38(6), 2019.

\bibitem{novotny2019perspectivenet}
David Novotny, Ben Graham, and Jeremy Reizenstein.
\newblock Perspectivenet: A scene-consistent image generator for new view
  synthesis in real indoor environments.
\newblock In {\em Advances in Neural Information Processing Systems}, 2019.

\bibitem{paul2018wtalc}
Sujoy Paul, Sourya Roy, and Amit~K. Roy-Chowdhury.
\newblock W-talc: Weakly-supervised temporal activity localization and
  classification.
\newblock In {\em ECCV}, 2018.

\bibitem{price2003preliminary}
Jill Price, Seago Catriona, Elizabeth~M Welsh, and Natalie~K Waran.
\newblock Preliminary evaluation of a behaviour--based system for assessment of
  post--operative pain in horses following arthroscopic surgery.
\newblock {\em Veterinary anaesthesia and analgesia}, 30(3), 2003.

\bibitem{pritchett2003identification}
Lori~C Pritchett, Catherine Ulibarri, Malcolm~C Roberts, Robert~K Schneider,
  and Debra~C Sellon.
\newblock Identification of potential physiological and behavioral indicators
  of postoperative pain in horses after exploratory celiotomy for colic.
\newblock {\em Applied Animal Behaviour Science}, 80(1), 2003.

\bibitem{raekallio1997comparison}
Marja Raekallio, Polly~M Taylor, and M Bloomfield.
\newblock A comparison of methods for evaluation of pain and distress after
  orthopaedic surgery in horses.
\newblock {\em Veterinary Anaesthesia and Analgesia}, 24(2), 1997.

\bibitem{rashid2018should}
M Rashid, S Broom{\'e}, PH Andersen, KB Gleerup, and YJ Lee.
\newblock What should i annotate? an automatic tool for finding video segments
  for equifacs annotation.
\newblock In {\em Measuring Behavior}, 2018.

\bibitem{rashid2017interspecies}
Maheen Rashid, Xiuye Gu, and Yong Jae~Lee.
\newblock Interspecies knowledge transfer for facial keypoint detection.
\newblock In {\em CVPR}, 2017.

\bibitem{rashid2020action}
Maheen Rashid, Hedvig Kjellstr{\"o}m, and Yong~Jae Lee.
\newblock Action graphs: Weakly-supervised action localization with graph
  convolution networks.
\newblock In {\em WACV}, 2020.

\bibitem{rashid2020equine}
Maheen Rashid, Alina Silventoinen, Karina~Bech Gleerup, and Pia~Haubro
  Andersen.
\newblock Equine facial action coding system for determination of pain-related
  facial responses in videos of horses.
\newblock {\em Plos one}, 15(11), 2020.

\bibitem{rhodin2019neural}
Helge Rhodin, Victor Constantin, Isinsu Katircioglu, Mathieu Salzmann, and
  Pascal Fua.
\newblock Neural scene decomposition for multi-person motion capture.
\newblock In {\em CVPR}, 2019.

\bibitem{rhodin2018unsupervised}
Helge Rhodin, Mathieu Salzmann, and Pascal Fua.
\newblock Unsupervised geometry-aware representation for {3D} human pose
  estimation.
\newblock In {\em ECCV}, 2018.

\bibitem{ronneberger2015u}
Olaf Ronneberger, Philipp Fischer, and Thomas Brox.
\newblock U-net: Convolutional networks for biomedical image segmentation.
\newblock In {\em International Conference on Medical image computing and
  computer-assisted intervention}. Springer, 2015.

\bibitem{sanakoyeu2020transferring}
Artsiom Sanakoyeu, Vasil Khalidov, Maureen~S. McCarthy, Andrea Vedaldi, and
  Natalia Neverova.
\newblock Transferring dense pose to proximal animal classes.
\newblock In {\em CVPR}, 2020.

\bibitem{sellon2004effects}
Debra~C Sellon, Malcolm~C Roberts, Anthony~T Blikslager, Catherine Ulibarri,
  and Mark~G Papich.
\newblock Effects of continuous rate intravenous infusion of butorphanol on
  physiologic and outcome variables in horses after celiotomy.
\newblock {\em Journal of Veterinary Internal Medicine}, 18(4), 2004.

\bibitem{shin20193d}
Daeyun Shin, Zhile Ren, Erik~B Sudderth, and Charless~C Fowlkes.
\newblock {3D} scene reconstruction with multi-layer depth and epipolar
  transformers.
\newblock In {\em ICCV}, 2019.

\bibitem{shu2017cvpr}
Zhixin Shu, Ersin Yumer, Sunil Hadap, Kalyan Sunkavalli, Eli Shechtman, and
  Dimitris Samaras.
\newblock Neural face editing with intrinsic image disentangling.
\newblock In {\em CVPR}, 2017.

\bibitem{singh2019finegan}
Krishna~Kumar Singh, Utkarsh Ojha, and Yong~Jae Lee.
\newblock Finegan: Unsupervised hierarchical disentanglement for fine-grained
  object generation and discovery.
\newblock In {\em CVPR}, 2019.

\bibitem{sitzmann2019deepvoxels}
Vincent Sitzmann, Justus Thies, Felix Heide, Matthias Nie{\ss}ner, Gordon
  Wetzstein, and Michael Zollhofer.
\newblock Deepvoxels: Learning persistent {3D} feature embeddings.
\newblock In {\em CVPR}, 2019.

\bibitem{sitzmann2019scene}
Vincent Sitzmann, Michael Zollh{\"o}fer, and Gordon Wetzstein.
\newblock Scene representation networks: Continuous {3D}-structure-aware neural
  scene representations.
\newblock In {\em Advances in Neural Information Processing Systems}, 2019.

\bibitem{tenenbaum2000separating}
Joshua~B Tenenbaum and William~T Freeman.
\newblock Separating style and content with bilinear models.
\newblock {\em Neural computation}, 12(6), 2000.

\bibitem{torcivia2021equine}
Catherine Torcivia and Sue McDonnell.
\newblock Equine discomfort ethogram.
\newblock {\em Animals}, 11(2), 2021.

\bibitem{tran2017disentangled}
Luan Tran, Xi Yin, and Xiaoming Liu.
\newblock Disentangled representation learning gan for pose-invariant face
  recognition.
\newblock In {\em CVPR}, 2017.

\bibitem{tulsiani2018factoring}
Shubham Tulsiani, Saurabh Gupta, David~F Fouhey, Alexei~A Efros, and Jitendra
  Malik.
\newblock Factoring shape, pose, and layout from the {2d} image of a {3D}
  scene.
\newblock In {\em CVPR}, 2018.

\bibitem{tuttle2018deep}
Alexander~H Tuttle, Mark~J Molinaro, Jasmine~F Jethwa, Susana~G Sotocinal,
  Juan~C Prieto, Martin~A Styner, Jeffrey~S Mogil, and Mark~J Zylka.
\newblock A deep neural network to assess spontaneous pain from mouse facial
  expressions.
\newblock {\em Molecular pain}, 14, 2018.

\bibitem{van2015monitoring}
Johannes~PAM van Loon and Machteld~C Van~Dierendonck.
\newblock Monitoring acute equine visceral pain with the equine utrecht
  university scale for composite pain assessment ({EQUUS-COMPASS}) and the
  equine utrecht university scale for facial assessment of pain ({EQUUS-FAP}):
  a scale-construction study.
\newblock {\em The Veterinary Journal}, 206(3), 2015.

\bibitem{wang2017untrimmednets}
Limin Wang, Yuanjun Xiong, Dahua Lin, and Luc Van~Gool.
\newblock Untrimmednets for weakly supervised action recognition and detection.
\newblock In {\em CVPR}, 2017.

\bibitem{wathan2015equifacs}
Jen Wathan, Anne~M Burrows, Bridget~M Waller, and Karen McComb.
\newblock {EquiFACS: The Equine Facial Action Coding System}.
\newblock {\em PLoS one}, 10(8), 2015.

\bibitem{wiles2020synsin}
Olivia Wiles, Georgia Gkioxari, Richard Szeliski, and Justin Johnson.
\newblock Synsin: End-to-end view synthesis from a single image.
\newblock In {\em CVPR}, 2020.

\bibitem{yang2016human}
Heng Yang, Renqiao Zhang, and Peter Robinson.
\newblock Human and sheep facial landmarks localisation by triplet interpolated
  features.
\newblock In {\em WACV}, 2016.

\bibitem{zuffi2019three}
Silvia Zuffi, Angjoo Kanazawa, Tanya Berger-Wolf, and Michael Black.
\newblock Three-d safari: Learning to estimate zebra pose, shape, and texture
  from images “in the wild”.
\newblock In {\em ICCV}, 2019.

\bibitem{zuffi2018lions}
Silvia Zuffi, Angjoo Kanazawa, and Michael~J Black.
\newblock Lions and tigers and bears: Capturing non-rigid, {3D}, articulated
  shape from images.
\newblock In {\em CVPR}, 2018.

\bibitem{zuffi20173d}
Silvia Zuffi, Angjoo Kanazawa, David~W Jacobs, and Michael~J Black.
\newblock {3D} menagerie: Modeling the {3D} shape and pose of animals.
\newblock In {\em CVPR}, 2017.

\end{thebibliography}
}

\appendix
\section{Further implementation details}
\begin{table}[t]
    \resizebox{\linewidth}{!}{%
        \centering
    	\begin{tabular}{c|c|c|c|c }
    	    \hline
        	Horse & \# No-Pain & \# Pain & \%age No-Pain & \%age Pain\\
            \hline
            Aslan & 414 & 376 & 52.41\% & 47.59\%\\
            Brava & 335 & 436 & 43.45\% & 56.55\%\\
            Herrera & 470 & 567 & 45.32\% & 54.68\%\\
            Inkasso & 330 & 439 & 42.91\% & 57.09\%\\
            Julia & 465 & 433 & 51.78\% & 48.22\%\\
            Kastanjett & 413 & 357 & 53.64\% & 46.36\%\\
            Naughty but Nice & 405 & 394 & 50.69\% & 49.31\%\\
            Sir Holger & 351 & 308 & 53.26\% & 46.74\%\\
            \hline
            Total & 3183 & 3310 & - & -\\
            \hline
	    \end{tabular}}
\caption{Number of pain and no-pain video segments per horse.}
\label{table_dset_details_video} 
\end{table}
\begin{table}[t]
    \resizebox{\linewidth}{!}{%
    \centering
    	\begin{tabular}{c|c|c|c|c }
    	    \hline
        	Horse & \# No-Pain & \# Pain & \%age No-Pain & \%age Pain\\
            \hline
            Aslan & 50974 & 53396 & 48.84\% & 51.16\%\\
            Brava & 54355 & 49651 & 52.26\% & 47.74\%\\
            Herrera & 46194 & 51964 & 47.06\% & 52.94\%\\
            Inkasso & 55805 & 53948 & 50.85\% & 49.15\%\\
            Julia & 48695 & 50902 & 48.89\% & 51.11\%\\
            Kastanjett & 55028 & 55990 & 49.57\% & 50.43\%\\
            Naughty but Nice & 54571 & 51117 & 51.63\% & 48.37\%\\
            Sir Holger & 55617 & 56129 & 49.77\% & 50.23\%\\
            \hline
            Total & 421239 & 423097 & - & -\\
            \hline
	    \end{tabular}}
\caption{Number of pain and no-pain frames per horse.}
\label{table_dset_details_frames} 
\end{table}
\begin{figure}[ht]
    \centering
    \includegraphics[width = \linewidth]{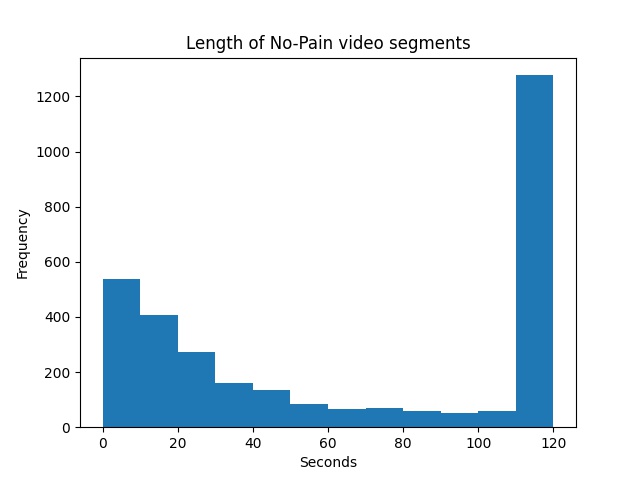}
    \includegraphics[width = \linewidth]{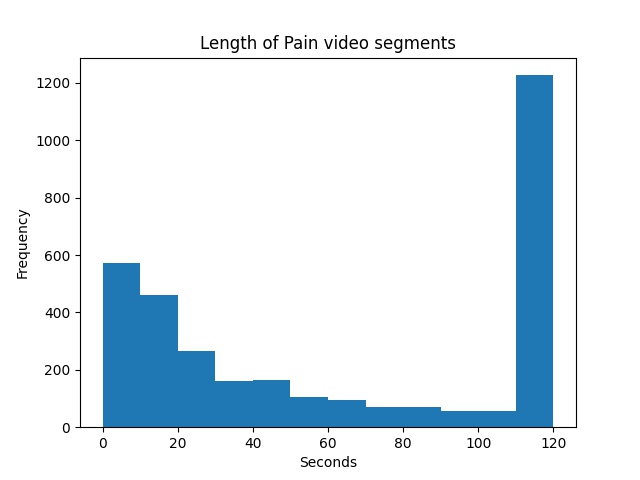}
    \caption{Histogram of video segments' length for pain and no-pain. Most data points have length 2 minutes.}
    \label{figure_length_hist}
\end{figure}

The dataset presented many practical challenges in terms of preprocessing. Videos from each camera were manually offset when necessary to sync temporally with other cameras in the stall. Time periods with humans present in the stall or corridor were manually marked for exclusion when not recorded in the experiment log. Technical faults led to intermittent recording from some cameras. Only time periods with footage from all cameras were used.

The cameras were calibrated to recover their intrinsic and extrinsic parameters by use of a large checkerboard pattern of known dimensions, solving Perspective-and-Point (PnP) problem using RANSAC in OpenCV~\cite{opencv_library}, and bundle adjustment.

Table \ref{table_dset_details_video} provides the number of video segments for each horse used for pain classification. Table~\ref{table_dset_details_frames} provides the same information information in terms of frames. The corresponding pain and no-pain percentages varies between these two tables since the segment length is variable.

We used `Naughty but Nice' as our validation horse, as it has the most balanced class distribution for video segments (Table~\ref{table_dset_details_video}). When testing on `Naughty but Nice', we used the first horse, `Aslan', as our validation subject. To keep results comparable we used the same subjects for validation when training on frames, even though the class distribution is different (Table~\ref{table_dset_details_frames}).

In Figure~\ref{figure_length_hist} we show the distribution of video segments' length in seconds. Most segments are two minutes in length. A slightly larger proportion of no-pain videos have 2 minutes length compared to their proportion in pain videos. This may be because horses display restlessness when in pain which makes their consistent detection in every frame more difficult. 

When detecting horsing with MaskRCNN~\cite{he2017mask}, we noticed high confusion between `horse' and `cow' categories, and included high confidence detections from both categories.

To compensate for unbalanced training data we use a weighted cross-entropy loss. The weights for each class are calculated as follows, where $p$ and $np$ are the number of training data points for pain and no-pain respectively:
\begin{equation}
\begin{aligned}
    w_{pain} = 2 (1-\frac{p}{p+np}),\\
    w_{no-pain} = 2 (1-\frac{np}{p+np}).
    \end{aligned}
\end{equation}
Multiplication by $2$ -- the number of classes -- keeps the weight around 1, hence maintaining the overall magnitude of the loss.

The maximum length of each video segment is 2 minutes. At 2 fps this equals 240 frames. The minimum length is 5 seconds (10 frames). The frame based model was trained at 1 fps. This was done to avoid very repetitive frames, and to speed up training.

The model trained from scratch was trained using a different schedule as the entire network had to be learned, and not just the pain classification head. We trained the model at 0.0001 learning rate for 50 epochs. We evaluated on validation and testing data after every 5 epochs, and used the results to determine both the `True' and `Oracle' performance.

The code base was is available at \url{https://github.com/menorashid/gross_pain}. The EOP dataset can be accessed by agreement for collaborative research with the authors of ~\cite{ask2020identification}.

\section{Qualitative examples}
\begin{figure*}
    \centering
    \includegraphics{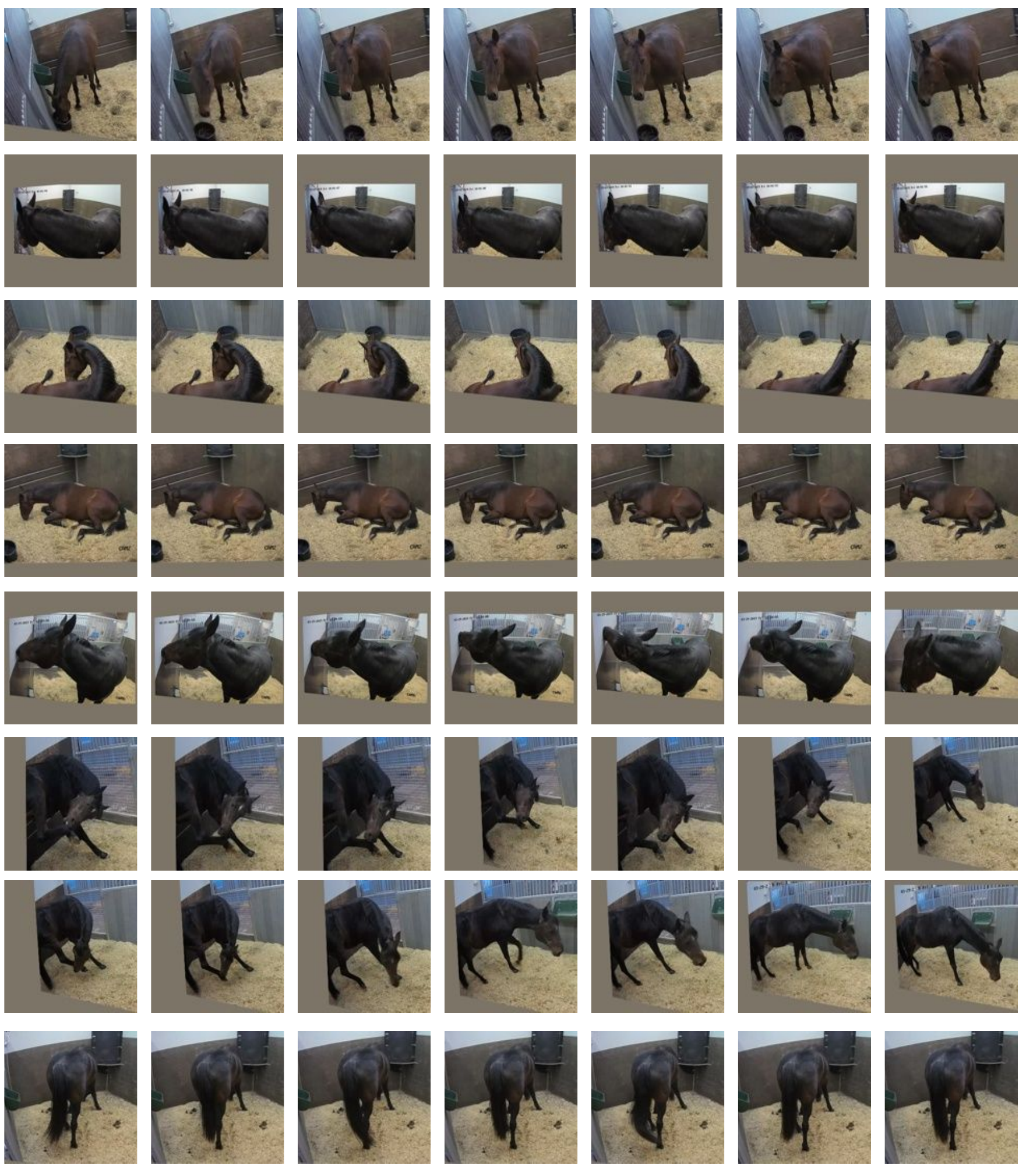}
    \caption{Painful behavior correctly classified by our network}
    \label{figure_pain_qual}
\end{figure*}
Figure~\ref{figure_pain_qual} shows further qualitative examples of video segments correctly classified as painful. 

Rows 1-2 show `lowered ears'~\cite{gleerup2015equine}, rows 3-4 show `lying down'~\cite{pritchett2003identification}, and row 5 shows stretching~\cite{gleerup2016recognition} similar to results in the main paper. In addition, we observe `looking at painful area', and `lowered head' in rows 6-7~\cite{gleerup2015equine}, and `frequent tail flicking' in row 8~\cite{van2015monitoring}.
`lowered ears'~\cite{gleerup2015equine} (second row), a lifted left hind limb (first row), corresponding to 
`non-weight bearing'~\cite{bussieres2008development}, `lying down' (third row), `looking at flank' (fourth row) 
~\cite{pritchett2003identification}, and one example of gross pain behavior, `stretching' (last row)~\cite{gleerup2016recognition}.

\end{document}